\documentclass[10pt,twocolumn,letterpaper]{article}

\usepackage{cvpr}
\usepackage{times}
\usepackage{epsfig}
\usepackage{graphicx}
\usepackage{amsmath}
\usepackage{amssymb}

\usepackage{enumerate}
\usepackage{multirow}

\usepackage{enumitem}
\usepackage{makecell}
\usepackage{tabulary}
\usepackage{pifont}
\usepackage{url}
\usepackage{subfigure}
\usepackage{threeparttable}

\usepackage[breaklinks=true,bookmarks=false]{hyperref}

\cvprfinalcopy 

\def\cvprPaperID{8334} 

\ifcvprfinal\fi
\begin{document}

\title{Exploring Category-Agnostic Clusters for Open-Set Domain Adaptation}

\author{Yingwei Pan$^{\dag}$, Ting Yao$^{\dag}$, Yehao Li$^{\dag}$, Chong-Wah Ngo$^{\ddag}$, and Tao Mei$^{\dag}$ \\
{\normalsize\centering$^{\dag}$ JD AI Research, Beijing, China}\\
{\normalsize\centering$^{\ddag}$ City University of Hong Kong, Kowloon, Hong Kong}\\
{\tt\small \{panyw.ustc, tingyao.ustc, yehaoli.sysu\}@gmail.com, cscwngo@cityu.edu.hk, tmei@jd.com}
}

\maketitle
\thispagestyle{empty}

\begin{abstract}
Unsupervised domain adaptation has received significant attention in recent years. Most of existing works tackle the closed-set scenario, assuming that the source and target domains share the exactly same categories. In practice, nevertheless, a target domain often contains samples of classes unseen in source domain (i.e., unknown class). The extension of domain adaptation from closed-set to such open-set situation is not trivial since the target samples in unknown class are not expected to align with the source. In this paper, we address this problem by augmenting the state-of-the-art domain adaptation technique, Self-Ensembling, with category-agnostic clusters in target domain. Specifically, we present Self-Ensembling with Category-agnostic Clusters (SE-CC) --- a novel architecture that steers domain adaptation with the additional guidance of category-agnostic clusters that are specific to target domain. These clustering information provides domain-specific visual cues, facilitating the generalization of Self-Ensembling for both closed-set and open-set scenarios. Technically, clustering is firstly performed over all the unlabeled target samples to obtain the category-agnostic clusters, which reveal the underlying data space structure peculiar to target domain. A clustering branch is capitalized on to ensure that the learnt representation preserves such underlying structure by matching the estimated assignment distribution over clusters to the inherent cluster distribution for each target sample. Furthermore, SE-CC enhances the learnt representation with mutual information maximization. Extensive experiments are conducted on Office and VisDA datasets for both open-set and closed-set domain adaptation, and superior results are reported when comparing to the state-of-the-art approaches.
\end{abstract}

\section{Introduction}

\begin{figure}[tbh]
\begin{center}
\vspace{-0.15in}
\includegraphics[width=0.92\linewidth]{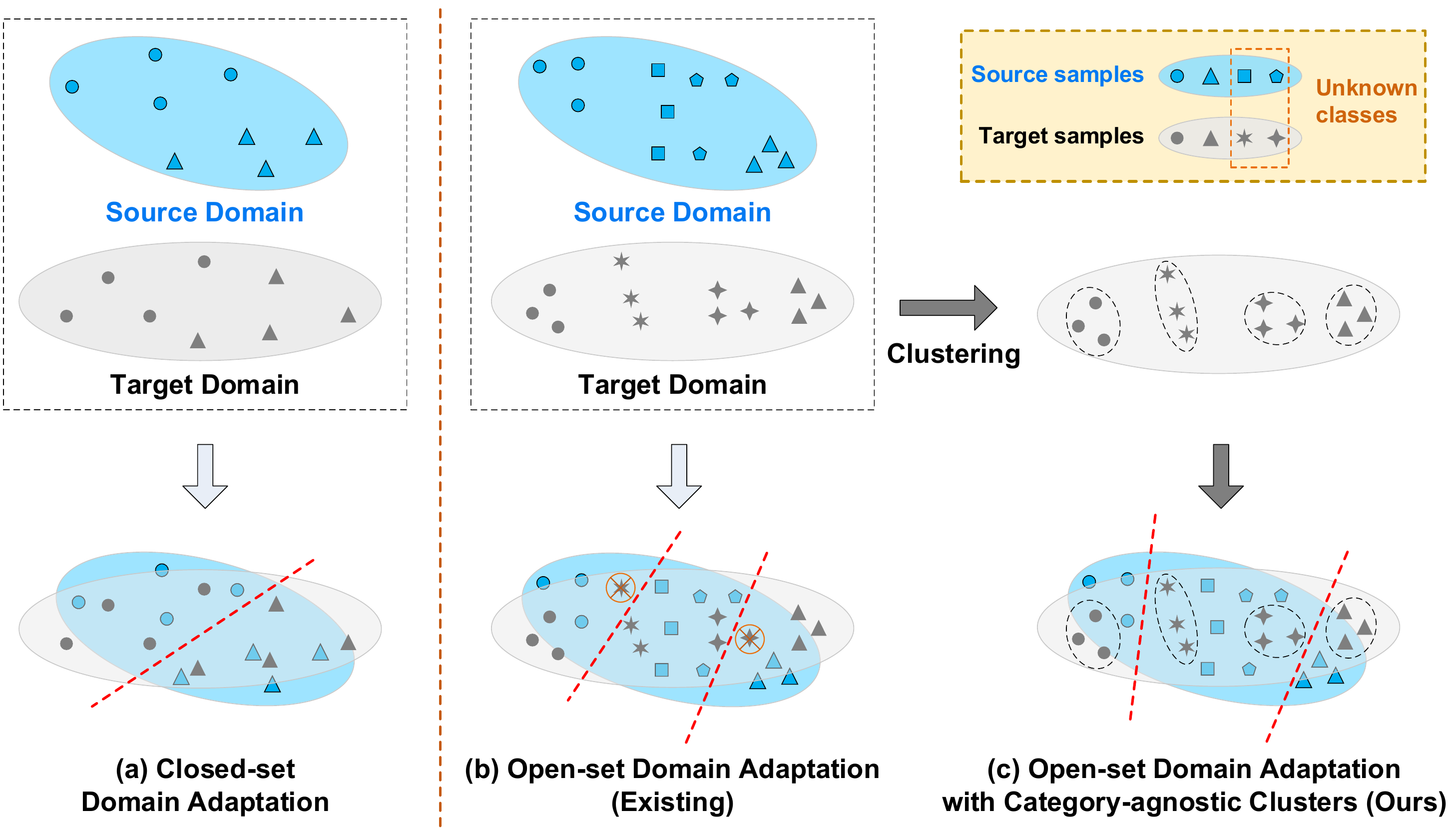}
\end{center}
\vspace{-0.2in}
\caption{\small A comparison between (a) closed-set domain adaptation, (b) existing methods for open-set domain adaptation, and (c) our open-set domain adaptation with category-agnostic clusters.}
\label{fig:fig1}
\vspace{-0.3in}
\end{figure}

Convolutional Neural Networks (CNNs) have driven vision technologies to reach new state-of-the-arts. The achievements, nevertheless, are on the assumption that large quantities of annotated data are accessible for model training. The assumption becomes impractical when cost-expensive and labor-intensive manual labeling is required. An alternative is to recycle off-the-shelf learnt knowledge/models in source domain for new domain(s). Unfortunately, the performance often drops significantly on a new domain, a phenomenon known as ``domain shift." One feasible way to alleviate this problem is to capitalize on unsupervised domain adaptation \cite{cai2019exploring,ganin2015unsupervised,long2016unsupervised,pan2019transferrable,yao2015semi,Zhang_2018_CVPR}, which leverages labeled source samples and unlabeled target samples to generalize a target model. One of the most critical limitations is that most existing models simply align data distributions between source and target domains. As a consequence, these models are only applicable in closed-set scenario (Figure \ref{fig:fig1}(a)) under the unrealistic assumption that both domains should share exactly the same set of categories. This adversely hinders the generalization of these models in open-set scenario to distinguish target samples of unknown class (unseen in source domain) from the target samples of known classes (seen in source~domain).

The difficulty of open-set domain adaptation mainly originates from two aspects: 1) how to distinguish the unknown target samples from known ones while classifying the known target samples correctly? 2) how to learn a hybrid network for both closed-set and open-set domain adaptation? One straightforward way (Figure \ref{fig:fig1}(b)) to alleviate the first issue is by employing an additional binary classifier for assigning known/unknown label to each target sample \cite{panareda2017open}. All the unknown target samples are further taken as outlier and will be discarded during the adaptation from source to target. As the unknown target samples are holistically grouped as one generic class, the inherent data structure is not fully exploited. In the case when the distribution of these target samples is diverse or the semantic labels between known and unknown classes are ambiguous, the performance of binary classification is suboptimal. Instead, we novelly perform clustering over all unlabeled target samples to explicitly model the diverse semantics of both known and unknown classes in target domain, as depicted in Figure \ref{fig:fig1}(c). All target samples are firstly decomposed into clusters, and the learnt clusters, though category-agnostic, convey the discriminative knowledge of unknown and known classes specific to target domain. As such, by further steering domain adaptation with category-agnostic clusters, the learnt representations are expected to be domain-invariant for known classes, and discriminative for unknown and known classes in target domain. To address the second issue, we remould Self-Ensembling \cite{french2018self} with an additional clustering branch to estimate the assignment distribution over all clusters for each target sample, which in turn refines the learnt representations to preserve inherent structure of target domain.

\begin{figure*}[!tb]
\vspace{-0.1in}
\centering {\includegraphics[width=0.98\textwidth]{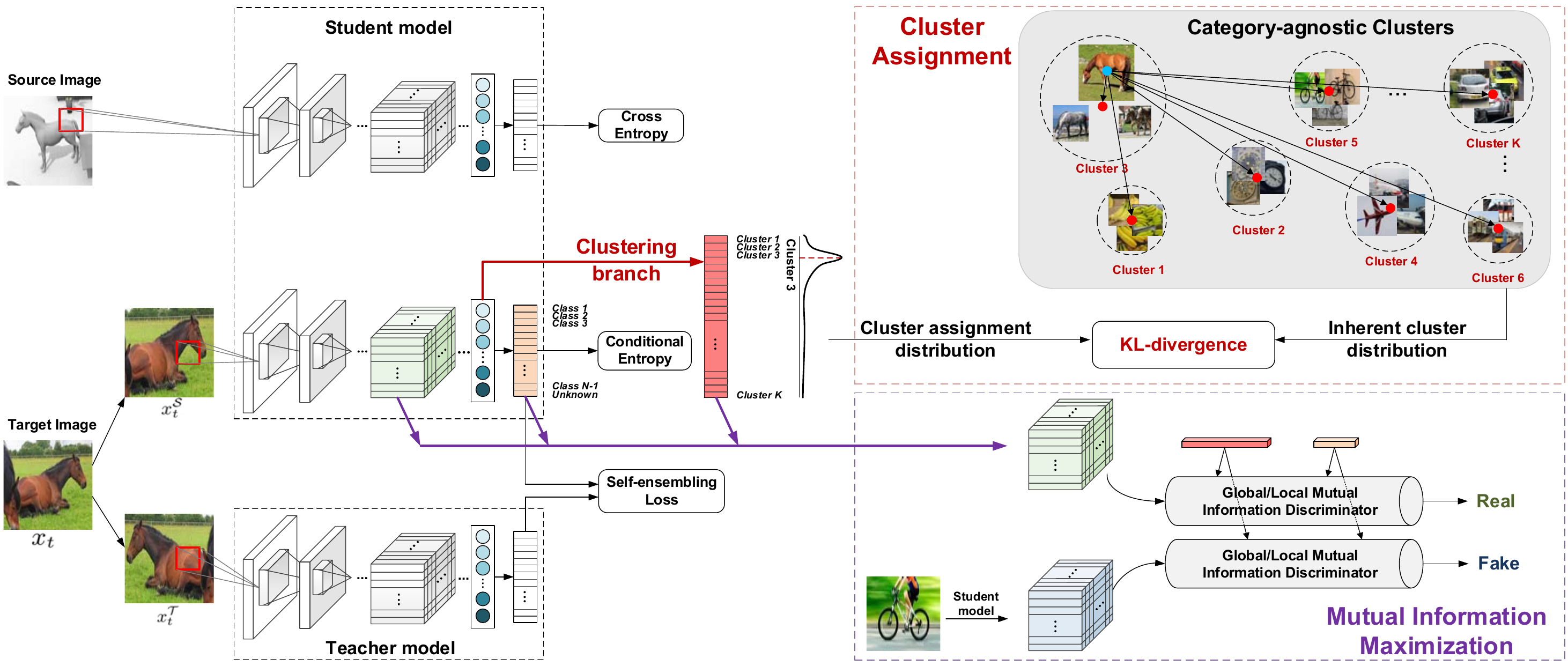}}
\caption{\small An overview of our SE-CC. Each labeled source image is fed into student model to train the classifier with cross entropy. Each unlabeled target image $x_t$ is transformed into two perturbed samples, i.e., $x_t^{\mathcal{S}}$ and $x_t^{\mathcal{T}}$, before injected into student and teacher models separately. Conditional entropy is applied to $x_t^{\mathcal{S}}$ in student pathway and self-ensembling loss is adopted to align the classification predictions between teacher and student. To further exploit the underlying data structure of target domain, we perform clustering to decompose the whole unlabeled target samples into a set of category-agnostic clusters (\textbf{top right}), which will be incorporated into Self-Ensembling to facilitate both closed-set and open-set scenarios. Specifically, an additional clustering branch is integrated into student to infer the assignment distribution over all clusters for each target sample $x_t^{\mathcal{S}}$. By aligning the estimated cluster assignment distribution to the inherent cluster distribution learnt from original clusters via minimizing their KL-divergence, the feature representation is enforced to preserve the underlying data structure in target domain. Furthermore, the feature representation of student is enhanced by maximizing the mutual information among its feature map, classification and cluster assignment distributions (\textbf{bottom right}). The maximization is conducted at both global and local levels as detailed in Figure \ref{fig:MIM}.}
\label{fig:2}
\vspace{-0.2in}
\end{figure*}

To this end, we present a new Self-Ensembling with Category-agnostic Clusters (SE-CC), as shown in Figure \ref{fig:2}. Specifically, clustering is firstly implemented to decompose all the target samples into a set of category-agnostic clusters. The underlying structure of each target sample is thus formulated as its inherent cluster distribution over all clusters, which is initially obtained by utilizing a softmax over the cosine similarities between this sample and each cluster centroid. With this, an additional clustering branch is integrated into student model of Self-Ensembling to predict the cluster assignment distribution of each target sample. For each target sample, the KL-divergence is exploited to model the mismatch between its estimated cluster assignment distribution and the inherent cluster distribution. By minimizing the KL-divergence, the learnt feature is enforced to preserve the underlying data structure in target domain. Moreover, we uniquely maximize the mutual information among the input intermediate feature map, the output classification distribution and cluster assignment distribution of target sample in student to further enhance the learnt feature representation. The whole SE-CC framework is jointly optimized.

\section{Related Work}
\textbf{Unsupervised Domain Adaptation.} One common solution for unsupervised domain adaptation in closed-set scenario is to learn transferrable feature in CNNs by minimizing domain discrepancy through Maximum Mean Discrepancy (MMD) \cite{gretton2012kernel}. \cite{tzeng2014deep} is one of early works that integrates MMD into CNNs to learn domain invariant representation. \cite{long2016unsupervised} additionally incorporates a residual transfer module into the MMD-based adaptation of classifiers. Inspired by \cite{goodfellow2014generative}, another direction of unsupervised domain adaptation is to encourage domain confusion across different domains via a domain discriminator \cite{chen2019mocycle,ganin2015unsupervised,tzeng2015simultaneous}, which is devised to predict the domain (source/target) of each input sample. In particular, a domain confusion loss \cite{tzeng2015simultaneous} in domain discriminator is devised to enforce the learnt representation to be domain invariant. \cite{ganin2015unsupervised} formulates domain confusion as a task of binary classification and utilizes a gradient reversal algorithm to optimize domain discriminator.

\textbf{Open-Set Domain Adaptation.} The task of open-set domain adaptation goes beyond the traditional domain adaptation to tackle a realistic open-set scenario, in which the target domain includes numerous samples from completely new and unknown classes not present in source domain. \cite{panareda2017open} is one of the early attempts to tackle the realistic open-set scenario. Busto \emph{et al.} additionally exploit the assignments of target samples as know/unknown classes when learning the mapping of known classes from source to target domain. Later on, \cite{saito2018open} utilizes adversarial training to learn feature representations that could separate the target samples of unknown class from the known target samples. Furthermore, \cite{baktashmotlagh2018learning} factorizes the source and target data into the shared and private subspace. The shared subspace models the target and source samples from known classes, while the target samples from unknown class are modeled with a private subspace, tailored to the target domain.

\textbf{Summary.} In summary, similar in spirit as previous methods \cite{baktashmotlagh2018learning,panareda2017open}, SE-CC utilizes unlabeled target samples for learning task-specific classifiers in the open-set scenario. Different from these approaches, SE-CC leverages category-agnostic clusters for representation learning. The learnt feature is driven to preserve the target data structure during domain adaption. The structure preservation enables effective alignment of sample distributions within known and unknown classes, and discrimination of samples between known and unknown classes. As a by-product, the preservation, which is represented as a cluster probability distribution, is exploited to further enhance representation learning. This is achieved through maximizing the mutual information among input feature, its cluster and class probability distributions. To the best of our knowledge, there is no study yet to fully explore the advantages of category-agnostic clusters for open-set domain adaptation.

\section{Our Approach: SE-CC}

In this paper, we remold Self-Ensembling to suit both closed-set and open-set scenarios by integrating category-agnostic clusters into domain adaptation procedure. An overview of our Self-Ensembling with Category-agnostic Clusters (SE-CC) model is depicted in Figure \ref{fig:2}.

\subsection{Notation}
In open-set domain adaptation, we are given the labeled samples $\mathcal{X}_s = \{(x_{s}, y_{s})\}$ in source domain and the unlabeled samples $\mathcal{X}_t = \{x_{t}\}$ in target domain belonging to $N$ classes, where $y_{s}$ is the class label of sample $x_{s}$. The set of $N$ classes is denoted as $\mathcal{C}$, which consists of $N-1$ known classes shared between two domains and an additional unknown class that aggregates all samples of unlabeled classes. The goal of open-set domain adaptation is to learn the domain-invariant representations and classifiers for recognizing the $N-1$ known classes in target domain and meanwhile distinguishing the unknown target samples from known ones.

\subsection{Self-Ensembling in Closed-Set Adaptation}
We first briefly recall the method of Self-Ensembling \cite{french2018self}. Self-Ensembling mainly builds upon the Mean Teacher \cite{tarvainen2017mean} for semi-supervised learning, which consists of a student model and a teacher model with the same network architecture. The main idea behind Self-Ensembling is to encourage consistent classification predictions between teacher and student under small perturbations of the input image. In other words, despite of different augmentations imposed on a target sample, both teacher and student models should predict similar classification probability distribution over all classes. Specifically, given two perturbed target samples $x_t^{\mathcal{S}}$ and $x_t^{\mathcal{T}}$ augmented from an unlabeled sample $x_t$, the self-ensembling loss penalizes the difference between the classification predictions of student and teacher:
\begin{equation}\label{eq:Preliminaries:Mean:eq-2}\small
	\mathcal{L}_{SE} ( x_t ) = || {\bf{P}}_{cls}^\mathcal{S} (x_t^{\mathcal{S}}) - {\bf{P}}_{cls}^\mathcal{T} (x_t^{\mathcal{T}}) ||^2_{2},
\end{equation}
where ${\bf{P}}_{cls}^\mathcal{S} (x_t^{\mathcal{S}})\in {\mathbb{R}}^{N}$  and ${\bf{P}}_{cls}^\mathcal{T}(x_t^{\mathcal{T}})\in {\mathbb{R}}^{N}$ denote the predicted classification distribution over $N$ classes via the classification branch in student and teacher. During training, the student is trained using gradient descent, while the weights of the teacher are directly updated as the exponential moving average of the student weights. Inspired by \cite{shu2018dirt}, we additionally adopt the unsupervised conditional entropy loss to train the classification branch in student, aiming to drive the decision boundaries of the classifier far away from high-density regions in target domain.

Therefore, the overall training loss of Self-Ensembling is composed of supervised cross entropy loss ($\mathcal{L}_{CSE}$) on source data, self-ensembling loss ($\mathcal{L}_{SE}$) and conditional entropy loss ($\mathcal{L}_{CDE}$) of unlabeled target data:
\begin{equation}\label{eq:Preliminaries:Mean:eq-3}\scriptsize
	\mathcal{L}_{SEC} =\sum_{ (x_{s}, y_{s}) \in \mathcal{S}} \mathcal{L}_{CSE} ( x_{s}, y_{s} )+ \sum_{x_t \in \mathcal{T} } (\mathcal{L}_{SE} (x_t)+\mathcal{L}_{CDE} (x_t)).
\end{equation}

\subsection{SE-CC for Open-Set Adaptation}
Open-set is more difficult than closed-set domain adaptation because it is required to classify not only inliers but also outliers into $N-1$ known and one unknown classes. The most typical way is by learning a binary classifier to recognize each target sample as known/unkown class. Nevertheless, such recipe oversimplifies the problem by assuming that all unknown samples belong to one class, while leaving the inherent data distribution among them unexploited. The robustness of this approach is questionable when the unknown samples span across multiple unknown classes and may not be properly grouped as one generic class. To alleviate this issue, we perform clustering to explicitly model the diverse semantics in target domain as the distilled category-agnostic clusters, which are further integrated into Self-Ensembling to guide domain adaptation. Specifically, we design an additional clustering branch in student of Self-Ensembling to align its estimated cluster assignment distribution with the inherent cluster distribution among category-agnostic clusters. Hence, the learnt feature representations are enforced to be domain-invariant for known classes and meanwhile more discriminative for unknown and known classes in target domain.

\textbf{Category-agnostic Clusters.} Clustering is an essential data analysis technique for grouping unlabeled data in unsupervised machine learning \cite{jain1999data}. Here we utilize $k$-means \cite{macqueen1967some}, the most popular clustering method, to decompose all unlabeled target samples $\mathcal{X}_t$ into a set of $K$ clusters $ \{C_{k}\}^{K}_{k=1}$, where $C_{k}$ represents the set of target samples from the $k$-th cluster. Accordingly, the obtained clusters $ \{C_{k}\}^{K}_{k=1}$, though category-agnostic, is still able to reveal the underlying structure tailored to target domain, where the target samples with similar semantics stay closer with local discrimination. In our implementations, we directly represent each target sample $x_{t}$ as the output feature (${\tilde{\bf{x}}_{t}}$) of CNNs pre-trained on ImageNet \cite{ILSVRC15} for clustering. We also tried to refresh the clusters according to learnt features periodically (e.g., every 5 training epoches), but that did not make a major difference.

We encode the underlying structure of each target sample $x_{t}$ as the joint relations between this sample and all category-agnostic clusters, i.e., the \emph{inherent cluster distribution} over all clusters. Specifically, for each target sample $x_{t}$, we measure its inherent cluster distribution $\tilde{\bf{P}}_{clu}( x_t ) \in {\mathbb{R}}^K$ through a softmax over the cosine similarities between this sample and each cluster centroid. The $k$-th element represents the cosine similarity between $x_{t}$ and the centroid $\mu_k$ of $k$-th cluster:
\begin{equation}\label{Eq:Eq1}\small
{\tilde{\bf{P}}^k_{clu}}( x_t ) = \frac{{{e^{ \rho\cdot {cos\left( {{\tilde{\bf{x}}_{t}},{\mu _k} } \right)} }}}}{{\sum\nolimits_{k'} {{e^{ \rho\cdot {cos\left( {{\tilde{\bf{x}}_{t}},{\mu _{k'}} } \right)} }}} }},~~{{\mu}_k}  = \frac{1}{{\left| {{C_k}} \right|}}\sum\limits_{{x_t} \in {C_k}} {\tilde{\bf{x}}_{t}},
\end{equation}
where $cos\left( \cdot \right)$ is cosine similarity function and $\rho$ is the temperature parameter of softmax for scaling. The centroid of each cluster $\mu_k$ is defined as the average of all samples belonging to that cluster.

\textbf{Clustering Branch.} An additional branch in student, named as \emph{clustering branch}, is especially designed to predict the distribution over all category-agnostic clusters for cluster assignment of each target sample $x_t^{\mathcal{S}}$. Concretely, we denote the feature of target sample $x_t^{\mathcal{S}}$ along student pathway as ${\bf{x}}_{t}^{\mathcal{S}} \in {\mathbb{R}}^M$. Hence, depending on the input feature ${\bf{x}}_{t}^{\mathcal{S}}$, clustering branch infers its \emph{cluster assignment distribution} ${\bf{P}}_{clu}( x_t^{\mathcal{S}} ) \in {\mathbb{R}}^K$ over all $K$ clusters via a modified softmax layer~\cite{liu2017sphereface}:
\begin{equation}\label{Eq:Eq2}\small
{{\bf{P}}^k_{clu}}( x_t^{\mathcal{S}} ) = \frac{{{e^{ \rho\cdot {cos\left( {{\bf{x}}_{t}^{\mathcal{S}},{\bf{W}}_{k} } \right)} }}}}{{\sum\nolimits_{k'} {{e^{ \rho\cdot {cos\left( {{\bf{x}}_{t}^{\mathcal{S}},{\bf{W}}_{k'} } \right)} }}} }},
\end{equation}
where ${{\bf{P}}^k_{clu}}( x_t^{\mathcal{S}} )$ is the $k$-th element in ${\bf{P}}_{clu}$ representing the probability of assigning target sample $x_t^{\mathcal{S}}$ into the $k$-th cluster. ${\bf{W}}_{k}$ is the $k$-th row of the parameter matrix ${\bf{W}} \in {\mathbb{R}}^{K\times M}$ in the modified softmax layer, which denotes the cluster assignment parameter matrix for the $k$-th cluster.

\textbf{KL-divergence Loss.} The clustering branch is trained with the supervision from the inherent cluster distribution of each target sample. To measure the mismatch between the estimated cluster assignment distribution and the inherent cluster distribution, a KL-divergence loss is defined as
\begin{equation}\label{Eq:Eq3}\small
\begin{array}{l}
\begin{split}
\mathcal{L}_{KL}= & \sum\limits_{x_t \in \mathcal{T} } {KL}\left( {\tilde{\bf{P}}_{clu}( x_t ) || {\bf{P}}_{clu}( x_t^{\mathcal{S}} )} \right) \\
= & \sum\limits_{x_t \in \mathcal{T} }\sum\nolimits_{k} {{{\tilde{\bf{P}}^k_{clu}}( x_t )}\log \left( {\frac{{{\tilde{\bf{P}}^k_{clu}}( x_t )}}{{{{\bf{P}}^k_{clu}}( x_t^{\mathcal{S}} )}}} \right)}.
\end{split}
\end{array}
\end{equation}
By minimizing the KL-divergence loss, the learnt representation is enforced to preserve the underlying data structure of target domain, pursuing to be more discriminative for both unknown and known classes. Moreover, we incorporate the inter-cluster relationship into the KL-divergence loss as a constraint to preserve the inherent relations among the cluster assignment parameter matrices. The spirit behind follows the philosophy that the cluster assignment parameter matrices of two semantically similar clusters should be similar. Hence, the KL-divergence loss with the constraint of inter-cluster relationships is formulated as
\begin{equation}\label{Eq:Eq4}\small
\begin{array}{l}
\mathcal{L}_{KL} = \sum\limits_{x_t \in \mathcal{T} } {KL}\left( {\tilde{\bf{P}}_{clu}( x_t ) || {\bf{P}}_{clu}( x_t^{\mathcal{S}} )} \right)\\
s.t.~~cos({\bf{W}}_{k},{\bf{W}}_{k'}) = cos({\mu _k},{\mu _{k'}}),1 \le k,k' \le K.
\end{array}
\end{equation}
The KL-divergence loss in Eq.(\ref{Eq:Eq4}) is further relaxed as:
\begin{equation}\label{Eq:Eq5}\small
\begin{array}{l}
\begin{split}
\mathcal{L}_{KL} = &\sum\limits_{x_t \in \mathcal{T} } {KL}\left( {\tilde{\bf{P}}_{clu}( x_t ) || {\bf{P}}_{clu}( x_t^{\mathcal{S}} )} \right) \\
& + \sum\limits_{1 \le k,k' \le K} | cos({\bf{W}}_{k},{\bf{W}}_{k'}) - cos({\mu _k},{\mu _{k'}})|.
\end{split}
\end{array}
\end{equation}

\subsection{Mutual Information Maximization in Student}
Given the input feature of a target sample, the student in our SE-CC produces both classification and cluster assignment distributions via the two parallel branches in a multi-task paradigm. To further strengthen the learnt target feature in an unsupervised manner, we leverage Mutual Information Maximization (MIM) \cite{hjelm2019learning} in student to maximize the mutual information among the input feature and the two output distributions. The rationale behind follows the philosophy that the global/local mutual information between input feature and output high-level features can be used to tune the feature's suitability for downstream tasks. As a result, we design a MIM module in student to simultaneously estimate and maximize the local and global mutual information among input feature map, the output classification distribution, and cluster assignment distribution.

\textbf{Global Mutual Information.} Technically, let ${\bf{x}}_t^{\mathcal{S}}\in {\mathbb{R}}^{H \times H \times D_0}$ be the output feature map of the last convolutional layer in student model for the input target sample $x_t^{\mathcal{S}}$ ($H$: the size of height and width; $D_0$: the number of channels). We encode this feature map into a global feature vector ${\bf{G}} ({\bf{x}}_t^{\mathcal{S}})\in {\mathbb{R}}^{D_1}$ via a convolutional layer (kernel size: $3 \times 3$; stride size: 1; filter number: $D_1$) plus an average pooling layer. Next, we concatenate the global feature vector ${\bf{G}} ({\bf{x}}_t^{\mathcal{S}})$ with the conditioning classification distribution ${\bf{P}}_{cls}^\mathcal{S} (x_t^{\mathcal{S}})$ and cluster assignment distribution ${\bf{P}}_{clu}( x_t^{\mathcal{S}} )$. The concatenated feature will be fed into the global Mutual information discriminator for discriminating whether the input global feature vector is aligned with the given classification and cluster assignment distributions. Here the global Mutual information discriminator is implemented with three stacked fully-connected network plus nonlinear activation. The final output score of global Mutual information discriminator is $V_g([{\bf{G}} ({\bf{x}}_t^{\mathcal{S}}),{\bf{P}}_{cls}^\mathcal{S} (x_t^{\mathcal{S}}),{\bf{P}}_{clu}( x_t^{\mathcal{S}} )])$, which represents the probability of discriminating the real input feature with matched classification and cluster assignment distributions. As such, the global Mutual Information is estimated via Jensen-Shannon MI estimator \cite{nowozin2016f}:
\begin{equation}\label{Eq:Eq6}\small
\begin{array}{l}
\mathcal{L}_g^{JSD}=\sum\limits_{x_t \in \mathcal{T} } -\varphi\left(-V_g([{\bf{G}} ({\bf{x}}_t^{\mathcal{S}}),{\bf{P}}_{cls}^\mathcal{S} (x_t^{\mathcal{S}}),{\bf{P}}_{clu}( x_t^{\mathcal{S}} )])\right)\\
~~~~~~- \sum\limits_{\hat {x}_t \in \mathcal{T},\hat {x}_t\neq x_t } \varphi\left(V_g([{\bf{G}} (\hat{\bf{x}}_t^{\mathcal{S}}),{\bf{P}}_{cls}^\mathcal{S} ( {x}_t^{\mathcal{S}}),{\bf{P}}_{clu}(  {x}_t^{\mathcal{S}} )])\right),
\end{array}
\end{equation}
where $\varphi\left( \cdot \right)$ is softplus function and ${\bf{G}} (\hat{\bf{x}}_t^{\mathcal{S}})$ denotes the global feature of a different target image $\hat {x}_t^{\mathcal{S}}$.

\begin{figure}[h]
\vspace{-0.20in}
\centering {\includegraphics[width=0.5\textwidth]{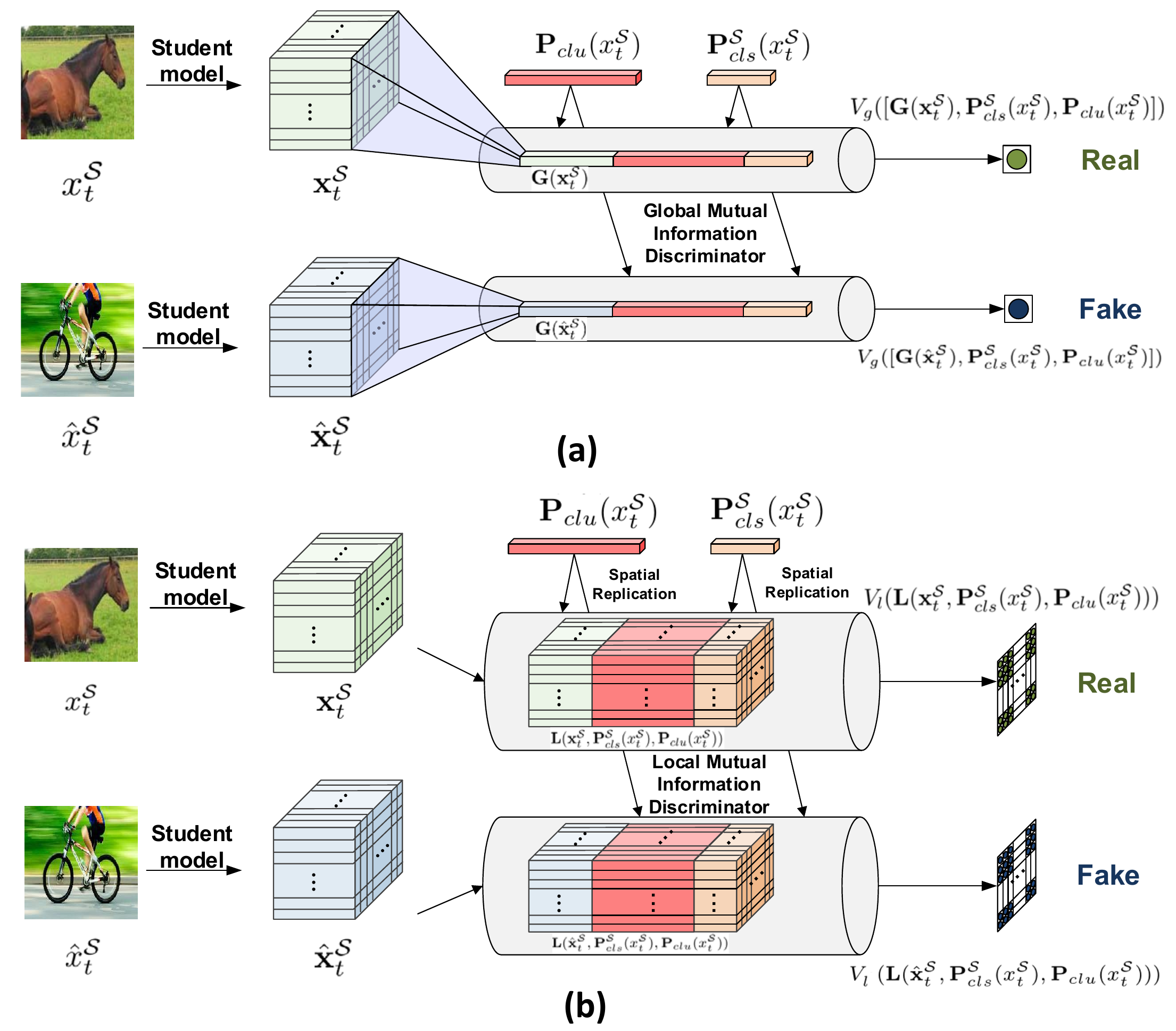}}
\vspace{-0.2in}
\caption{\small Framework of (a) global mutual information estimation and (b) local mutual information estimation in our SE-CC.}
\label{fig:MIM}
\vspace{-0.15in}
\end{figure}

\textbf{Local Mutual Information.} In addition, we exploit the local Mutual Information among the local input feature at every spatial location, and the output classification and cluster assignment distributions. In particular, we spatially replicate the two distributions ${\bf{P}}_{cls}^\mathcal{S} (x_t^{\mathcal{S}})$ and ${\bf{P}}_{clu}( x_t^{\mathcal{S}} )$ to construct $H \times H \times N$ and $H \times H \times K$ feature maps respectively, and then concatenate them with the input feature map ${\bf{x}}_t^{\mathcal{S}}$ along the channel dimension. The concatenated feature map ${\bf{L}}({\bf{x}}_t^{\mathcal{S}},{\bf{P}}_{cls}^\mathcal{S} (x_t^{\mathcal{S}}),{\bf{P}}_{clu}( x_t^{\mathcal{S}} ))\in {\mathbb{R}}^{H \times H \times (D_0+N+K)}$ will be fed into the local Mutual information discriminator for discriminating whether each input local feature is matched with the given classification and cluster assignment distributions. The local Mutual information discriminator is constructed with three stacked convolutional layer (kernel size: $1 \times 1$) plus nonlinear activation. Hence the final output score map of local Mutual information discriminator is $V_l({\bf{L}}({\bf{x}}_t^{\mathcal{S}},{\bf{P}}_{cls}^\mathcal{S} (x_t^{\mathcal{S}}),{\bf{P}}_{clu}( x_t^{\mathcal{S}} )))\in {\mathbb{R}}^{H \times H}$. The $i$-th element $V^{i}_l({\bf{L}}({\bf{x}}_t^{\mathcal{S}},{\bf{P}}_{cls}^\mathcal{S} (x_t^{\mathcal{S}}),{\bf{P}}_{clu}( x_t^{\mathcal{S}} )))$ in score map denotes the probability of discriminating the real input local feature at the $i$-th spatial location with matched classification and cluster assignment distributions. As such, the local Mutual Information is estimated as:
\begin{equation}\label{Eq:Eq7}\small
\begin{array}{l}
\mathcal{L}_l^{JSD}=\sum\limits_{x_t \in \mathcal{T} } -\frac{1}{{ {{H}^2} }}\sum\limits^{{H}^2}_{i=1}\varphi \left(-V^{i}_l({\bf{L}}({\bf{x}}_t^{\mathcal{S}},{\bf{P}}_{cls}^\mathcal{S} (x_t^{\mathcal{S}}),{\bf{P}}_{clu}( x_t^{\mathcal{S}} )))\right)\\
~~~~~~- \sum\limits_{\hat {x}_t \in \mathcal{T},\hat {x}_t\neq x_t } \frac{1}{{ {{H}^2} }}\sum\limits^{{H}^2}_{i=1}\varphi \left( V^{i}_l({\bf{L}}(\hat{\bf{x}}_t^{\mathcal{S}},{\bf{P}}_{cls}^\mathcal{S} ( x_t^{\mathcal{S}}),{\bf{P}}_{clu}( x_t^{\mathcal{S}} )))\right).
\end{array}
\end{equation}
Accordingly, the final objective for MIM module is measured as the combination of local and global Mutual Information estimations, balanced with tradeoff parameter $\alpha$:
\begin{equation}\label{Eq:Eq8}\small
\mathcal{L}_{MIM}= \alpha \mathcal{L}_g^{JSD}+\mathcal{L}_l^{JSD}.
\end{equation}
Figure \ref{fig:MIM} conceptually depicts the process of both local and global mutual information estimation.

\subsection{Training}
The overall training objective of our SE-CC integrates the cross entropy loss on source data, unsupervised self-ensembling loss, and conditional entropy loss in Eq.(\ref{eq:Preliminaries:Mean:eq-3}), KL-divergence loss of clustering branch in Eq.(\ref{Eq:Eq5}), and the Mutual Information estimation in Eq.(\ref{Eq:Eq8}) on target data:
\begin{equation}\label{Eq:Eq9}\small
\begin{array}{l}
\begin{split}
\mathcal{L}= \mathcal{L}_{SEC} +\mathcal{L}_{KL}-{\beta}\mathcal{L}_{MIM},
\end{split}
\end{array}
\end{equation}
where $\beta$ is tradeoff parameter.


\begin{table*}[!tb]\scriptsize
  \setlength\tabcolsep{6.5pt}
  \centering
  \vspace{-0.1in}
  \caption{\small Performance comparison with the state of arts on Office for open-set domain adaptation. $^\diamondsuit$ indicates a different open-set setting without unknown source examples.}
  \label{table:open-office}
  \begin{tabular}{l|cc|cc|cc|cc|cc|cc|cc}
  \Xhline{2\arrayrulewidth}
  \multirow{2}{*}{Method} & \multicolumn{2}{c|}{A $\rightarrow$ D} & \multicolumn{2}{c|}{A $\rightarrow$ W} & \multicolumn{2}{c|}{D $\rightarrow$ A} & \multicolumn{2}{c|}{D $\rightarrow$ W} & \multicolumn{2}{c|}{W $\rightarrow$ A} & \multicolumn{2}{c|}{W $\rightarrow$ D} & \multicolumn{2}{c}{Avg} \\
                    & OS   & OS*  & OS   & OS*  & OS   & OS*  & OS   & OS*  & OS   & OS*  & OS   & OS*  & OS   & OS*   \\\hline
  Source-only       & 67.1 & 67.0 & 64.6 & 63.8 & 61.9 & 60.7 & 90.6 & 92.3 & 60.2 & 59.7 & 96.7 & 98.7 & 73.5 & 73.7 \\
  RTN \cite{long2016unsupervised}             & 76.6 & 74.7 & 73.0 & 70.8 & 57.2 & 53.8 & 89.0 & 88.1 & 62.4 & 60.2 & 98.8 & 98.3 & 76.2 & 74.3  \\
  RevGrad \cite{ganin2015unsupervised}         & 78.3 & 77.3 & 75.9 & 73.8 & 57.6 & 54.1 & 89.8 & 88.9 & 64.0 & 61.8 & 98.7 & 98.0 & 77.4 & 75.7  \\
  AODA$^{\diamondsuit}$ \cite{saito2018open}            & 76.6 & 76.4 & 74.9 & 74.3 & 62.5 & 62.3 & 94.4 & 94.6 & 81.4 & 81.2 & 96.8 & 96.9 & 81.1 & 80.9  \\
  ATI-$\lambda$ \cite{panareda2017open}   & 79.8 & 79.2 & 77.6 & 76.5 & 71.3 & 70.0 & 93.5 & 93.2 & 76.7 & 76.5 & 98.3 & 99.2 & 82.9 & 82.4  \\
  FRODA \cite{baktashmotlagh2018learning}    & \textbf{88.0} & -    & 78.7 & -    & 76.5 & -    & \textbf{98.0} & -    & 73.7 & -    & 94.6 & -    & 84.9 & -     \\\hline
  SE-CC$^{\diamondsuit}$ &80.6&84.0 &82.4& 84.2& 83.2& \textbf{90.3} & 92.9& 96.6& 82.7& 85.9& 96.8& 99.1& 86.4& 90.0\\
  SE-CC               & 85.3 & \textbf{84.5} & \textbf{85.1} & \textbf{84.3} & \textbf{87.9} & 89.5 & 97.7 & \textbf{97.8} & \textbf{86.8} & \textbf{87.5} & \textbf{99.4} & \textbf{99.6} & \textbf{90.4} & \textbf{90.5}   \\
  \Xhline{2\arrayrulewidth}
  \end{tabular}
  \vspace{-0.1in}
\end{table*}

\begin{table*}[!tb]\small
  \centering
   \setlength\tabcolsep{3.5pt}
  \caption{\small Performance comparison with the state of arts on VisDA for open-set adaptation (Known-to-Unknown Ratio = 1:10). $^{ \diamondsuit}$ indicates a different open-set setting without unknown source examples. $^{\dag}$ indicates the results are referred from the official leaderboard \cite{VisDA2018}.}
  \begin{tabular}{l|ccccccccccccc|c|c|c}
  \Xhline{2\arrayrulewidth}
  Method        & aero  & bike  & bus   & car   & horse & knife & mbike & person & plant & skbrd & train & truck & unk   & Knwn  & Mean  & Overall  \\\hline
  Source-only   & 53.8  & 54.2  & 50.3  & 48.7  & 72.7  & 5.3   & 82.0  & 27.0   & 49.6  & 43.4  & 78.0  & 5.1   & 44.2  & 46.9  & 47.3  & 44.8 \\
  RevGrad \cite{ganin2015unsupervised}  & 33.0  & 57.3  & 44.1  & 33.9  & 72.1   & 46.9  & 82.2  & 26.8  & 36.8  & 50.4  & 89.4  & 9.8   & 47.8 & 48.6 & 48.5 & 47.8     \\
  RTN \cite{long2016unsupervised} & 49.2 & 72.6 & 66.5 & 39.5 & 80.8 & 18.8 & 73.8 & 56.8 & 47.4 & 45.2 & 74.0 & 4.5 & 48.7 & 52.4 & 52.1 & 49.0 \\
  SE$^{\dag}$ \cite{french2018self}    & \textbf{94.2}  & 74.1  & \textbf{86.1} & 68.1  & \textbf{91.0} & 26.1 & \textbf{95.2} & 46.0 & \textbf{85.0}  & 40.4    & 79.2    & 11.0    & 51.0  & 66.4  & 65.2  & 52.7       \\
  AODA$^{\diamondsuit}$$^{\dag}$ \cite{saito2018open}       & 80.2  & 63.1  & 59.1  & 63.1  & 83.2  & 12.1  & 89.1  & 5.0    & 61.0  & 14.0  & 79.2  & 0.0   & 69.0  & 50.8  & 52.2  & 67.6     \\
  ATI-$\lambda$ \cite{panareda2017open} & 85.7  & 74.9  & 60.3  & 49.9  & 80.0  & 19.3  & 88.8  & 40.8   & 54.0  & \textbf{59.2}  & 66.4  & \textbf{18.2}  & 59.5  & 58.1  & 58.2  & 59.3     \\ \hline
  SE-CC$^{\diamondsuit}$ &82.1 & 80.7 &59.7&50.0&80.6&36.7& 83.1&56.2& 56.6& 21.9& 57.7& 4.0& 70.6& 55.8& 56.9& 69.2\\
  SE-CC          & \textbf{94.2}  & \textbf{79.0}  & 83.4  & \textbf{70.7}  & \textbf{91.0}  & \textbf{43.5}  & 89.3  & \textbf{73.3}   & 69.4  & 58.8  & \textbf{79.4}  & 12.8  & \textbf{71.6}  & \textbf{70.4}  & \textbf{70.5}  & \textbf{71.6}     \\
  \Xhline{2\arrayrulewidth}
  \end{tabular}
  \vspace{-0.2in}
  \label{table:open-visda}
\end{table*}

\section{Experiments}

We empirically verify the merit of our SE-CC by conducting experiments on \emph{Office} \cite{saenko2010adapting} and \emph{VisDA} \cite{peng2018syn2real} datasets for both open-set and closed-set domain adaptation.

\textbf{Office} is the standard benchmark for domain adaptation, which contains 4,110 images from 31 categories. They are collected from three domains: Amazon (A), DSLR (D), and Webcam (W). Six directions of transfer among them are evaluated for both open-set and closed-set adaptation. For open-set adaptation, as in \cite{panareda2017open}, we firstly take 10 classes as the known classes shared between source and target domains. In alphabetical order, the classes with labels 11-20 are taken as the unknown classes in source, and the ones with labels 21-31 are unknown classes in target. Two metrics OS and OS*, are adopted for evaluation (OS: the accuracy on all known \& unknown target samples; OS*: the accuracy on the target samples of the 10 known classes). We adopt AlexNet \cite{krizhevsky2012imagenet} pre-trained on ImageNet \cite{ILSVRC15} as the basic CNNs architecture for clustering and adaptation. For closed-set adaptation, we follow \cite{long2017deep} and report accuracy on target domain over all 31 classes. The basic architecture of CNNs for clustering and adaptation is ResNet50 \cite{he2016resnet} pre-trained on ImageNet.

\textbf{VisDA} is a large-scale dataset for the challenging synthetic-real image transfer, consisting of 280k images from three domains. The synthetic images generated from 3D CAD models are taken as the training domain. The validation domain contains real images from COCO \cite{lin2014microsoft} and the testing domain includes video frames in YTBB \cite{real2017youtube}. Given the fact that the ground truth of testing set are not publicly available, the synthetic images in training domain are taken as source and the COCO images in validation domain are taken as target for evaluation. In particular, for open-set adaptation, we follow the open-set setting in \cite{peng2018syn2real} and take the 12 classes as the known classes for source \& target domains, the 33 background classes as the unknown classes in source, and the other 69 COCO categories as the unknown classes in target. The known-to-unknown ratio of samples in target domain is strictly set as 1:10. Three metrics, i.e., Knwn, Mean, and Overall, are adopted for evaluation. Here Knwn denotes the accuracy averaged over all known classes, Mean is the accuracy averaged over all known \& unknown classes, and Overall is the accuracy over all target samples. For closed-set adaptation, we report the accuracy of all the 12 classes for adaptation, as in the closed-set setting of \cite{peng2018syn2real}. We utilize ResNet152 as the backbone of CNNs for clustering and adaptation in both closed-set and open-set scenarios.

\begin{table*}[!tb]\small
  \centering
  \setlength\tabcolsep{6.5pt}
  \vspace{-0.1in}
  \caption{\small Performance comparison with the state of arts on VisDA dataset for closed-set domain adaptation.}
  \begin{tabular}{l|cccccccccccc|c}
  \Xhline{2\arrayrulewidth}
  Method & aero & bike & bus & car & horse & knife & mbike & person & plant & skbrd & train & truck & Mean \\ \hline
  Source-only & 67.1 & 51.4 & 50.8 & 64.5 & 83.4 & 13.0 & 89.9 & 34.4 & 78.8 & 47.0 & 88.1 & 2.0 & 55.9 \\
  RevGrad \cite{ganin2015unsupervised} & 81.9 & 77.7 & 82.8 & 44.3 & 81.2 & 29.5 & 65.1 & 28.6 & 51.9 & 54.6 & 82.8 & 7.8 & 57.4 \\
  RTN \cite{long2016unsupervised} & 89.1 & 56.4 & 72.4 & 69.7 & 77.9 & 49.5 & 87.7 & 13.0 & 88.1 & 77.4 & 86.7 & 7.2 & 64.6 \\
  MCD \cite{saito2018maximum} & 87.0 & 60.9 & 83.7 & 64.0 & 88.9 & 79.6 & 84.7 & 76.9 & 88.6 & 40.3 & 83.0 & 25.8 & 71.9 \\
  SimNet \cite{pinheiro2018unsupervised} & 94.3 & 82.3 & 73.5 & 47.2 & 87.9 & 49.2 & 75.1 & 79.7 & 85.3 & 68.5 & 81.1 & 50.3 & 72.9 \\
  TPN \cite{pan2019transferrable} & 93.7 & 85.1 & 69.2 & \textbf{81.6} & {93.5} & 61.9 & {89.3} & 81.4 & {93.5} & 81.6 & {84.5} & {49.9} & {80.4}\\
  SE \cite{french2018self} & 96.2 & \textbf{87.8} & \textbf{84.4} & 66.5 & \textbf{96.1} & 96.1 & 90.5 & 81.5 & \textbf{95.3} & 91.5 & 87.5 & 51.6 & 85.4 \\ \hline
  SE-CC & \textbf{96.3} & 86.5 & 82.4 & {81.3} & \textbf{96.1} & \textbf{97.2} & \textbf{91.2} & \textbf{84.7} & 94.4 & \textbf{94.1} & \textbf{88.3} & \textbf{53.4} & \textbf{87.2} \\
  \Xhline{2\arrayrulewidth}
  \end{tabular}
  \label{table:close-visda}
  \vspace{-0.20in}
\end{table*}

\textbf{Implementation Details.}
Our SE-CC is mainly implemented with PyTorch and the network weights are optimized with SGD. We set the learning rate and mini-batch size as 0.001 and 56 for all experiments. The maximum training iteration is set as 300 and 25 epochs on Office and VisDA, respectively. The dimension $D_1$ of global feature for global Mutual Information estimation is set as 128/1,024 in the backbone of AlexNet/ResNet. The number of clusters $K$ is determined using Gap statistics method ($K=25$ for Office and $K=500$ for VisDA). As in \cite{hjelm2019learning}, we restrict the hyper-parameter search for each dataset in range of $\alpha=\{1,5,10\}$ and $\beta=\{10^{-4},10^{-3},10^{-2}\}$ ($\alpha=1$, $\beta=10^{-3}$ for Office, and $\alpha=5$, $\beta=10^{-2}$ for VisDA).

\begin{table}[!tb]\scriptsize
  \centering
  \setlength\tabcolsep{1.5pt}
  \caption{\small Performance comparison with the state of arts on Office dataset for closed-set domain adaptation.}
  \begin{tabular}{l|c|c|c|c|c|c|c}
  \Xhline{2\arrayrulewidth}
  Method & A $\rightarrow$ D & A $\rightarrow$ W & D $\rightarrow$ A & D $\rightarrow$ W & W $\rightarrow$ A & W $\rightarrow$ D & Avg \\ \hline
  RTN \cite{long2016unsupervised} & 77.5 & 84.5 & 66.2 & 96.8 & 64.8 & 99.4 & 81.6 \\
  RevGrad \cite{ganin2015unsupervised} & 79.7 & 82.0 & 68.2 & 96.9 & 67.4 & 99.1 & 82.2 \\
  JAN \cite{long2017deep} & 85.1 & 86.0 & 69.2 & 96.7 & 70.7 & 99.7 & 84.6 \\
  SimNet \cite{pinheiro2018unsupervised} & 85.3 & 88.6 & 73.4 & 98.2 & 71.8 & 99.7 & 86.2 \\
  GTA \cite{sankaranarayanan2018generate} & 87.7 & 89.5 & 72.8 & 97.9 & 71.4 & 99.8 & 86.5 \\
  iCAN \cite{zhang2018collaborative} & 90.1 & \textbf{92.5} & 72.1 & 98.8 & 69.9 & \textbf{100} & 87.2 \\ \hline
  SE-CC & \textbf{91.4} & 90.7 & \textbf{74.0} & \textbf{99.0} & \textbf{72.9} & \textbf{100} & \textbf{88.0} \\
  \Xhline{2\arrayrulewidth}
  \end{tabular}
  \vspace{-0.25in}
  \label{table:close-office}
\end{table}

\subsection{Performance Comparison}

\textbf{Open-Set Adaptation on Office.} The results of different models on Office for open-set adaptation are shown in Table \ref{table:open-office}. It is worth noting that AODA adopts a
different open-set setting where unknown source samples are absent. For fair comparison with AODA, we additionally include a variant of our SE-CC (dubbed as SE-CC$^{ \diamondsuit}$) which learns classifier without unknown source samples. Specifically, the classifier in SE-CC$^{ \diamondsuit}$ is naturally able to recognize only the N-1 known classes and the target samples will be recognized as unknown if the predicted probability is lower than the threshold for any class as in open set SVM \cite{jain2014multi}.

Overall, the results across two metrics consistently indicate that our SE-CC obtains better performances against other state-of-the-art closed-set adaptation models (RTN and RevGrad) and open-set adaptation methods (AODA, ATI-$\lambda$, and FRODA) on most transfer directions. Please also note that our SE-CC improves the classification accuracy evidently on the harder transfers, e.g., D $\rightarrow$ A and W $\rightarrow$ A, where the two domains are substantially different. The results generally highlight the key advantage of exploiting underlying target data structure implicit in category-agnostic clusters for open-set domain adaptation. Such design makes the learnt feature representation to be domain-invariant for known classes while discriminative enough to segregate target samples from known and unknown classes. Specifically, by aligning the data distributions between source and target domains, RTN and RevGrad exhibit better performance than Source-only that trains classifier only on source data while leaving unlabeled target data unexploited. By rejecting unknown target samples as outliers and aligning data distributions only for inliers, the open-set adaptation techniques (AODA, ATI-$\lambda$, and FRODA)  outperform RTN and RevGrad. This confirms the effectiveness of excluding unknown target samples from the known target samples during domain adaptation in open-set scenario. Nevertheless, AODA, ATI-$\lambda$, and FRODA are still inferior to our SE-CC which steers the domain adaptation by injecting the distribution of category-agnostic clusters as a constraint for feature learning and alignment.

\textbf{Open-Set Adaptation on VisDA.} The performance comparison on VisDA for open-set adaptation is summarized in Table \ref{table:open-visda}. Our SE-CC performs consistently better than other methods across all the three metrics. In particular, the Mean accuracy averaged over 12 known classes plus one unknown class of our SE-CC can achieve 70.5\%, making the absolute improvement over the best closed-set adaptation method (SE) and open-set adaptation approach (ATI-$\lambda$) by 5.3\% and 12.3\%, respectively. Similar to the observations on Office for open-set adaptation, the open-set adaptation approaches (AODA and ATI-$\lambda$) exhibit better performance than RTN and RevGrad, by additionally separating unknown target samples from known target samples for open-set adaptation. Note that although the closed-set technique SE achieves higher Mean per-category accuracy than the open-set techniques (AODA and ATI-$\lambda$), the Overall accuracy over all target samples of SE are still worse than open-set techniques. This is because SE aligns unknown samples across different domains and thus fails to recognize unknown target samples. Furthermore, by integrating category-agnostic clusters into SE and steering domain adaptation to preserve the underlying target data structure of both known and unknown classes, SE-CC boosts the performances in terms of all metrics.

\textbf{Closed-Set Adaptation on Office and VisDA.} To further verify the generality of our proposed SE-CC, we additionally conduct experiments for domain adaptation in closed-set scenario. Tables \ref{table:close-office} and \ref{table:close-visda} show the performance comparisons on Office and VisDA datasets for closed-set domain adaptation. Similar to the observations for open-set domain adaptation task on these two datasets, our SE-CC achieves better performances than other state-of-the-art closed-set adaptation techniques. The results basically demonstrate the advantage of exploiting the underlying data structure in target domain via category-agnostic clusters, for domain adaptation, even on closed-set scenario without any diverse and ambiguous unknown samples.

\begin{table}[!tb]\small
\centering
  \vspace{-0.0in}
 \caption{\small Performance contribution of each design (i.e., Conditional Entropy (\textbf{CE}), KL-divergence Loss (\textbf{KL}), and Mutual Information Maximization (\textbf{MIM})) in SE-CC on VisDA for open-set transfer.}
  \begin{tabular}{l|ccc|ccc}
  \Xhline{2\arrayrulewidth}
        Method&    CE   & KL        & MIM        & Knwn    & Mean    & Overall \\ \hline
  SE & & &&66.4 & 65.2&52.7\\
  +CE&$\checkmark$   &               &                & 67.3    & 66.3    & 55.8   \\
  +KL&$\checkmark$   & $\checkmark$  &                & 69.3    & 69.3    & 69.1   \\\hline
  SE-CC&$\checkmark$   & $\checkmark$  & $\checkmark$   & \textbf{70.4}    & \textbf{70.5}    & \textbf{71.6}   \\
  \Xhline{2\arrayrulewidth}
  \end{tabular}

  \label{table:ablation}
  \vspace{-0.25in}
\end{table}

\begin{table}[!tb]\small
  \centering
  \vspace{-0.1in}
  \setlength\tabcolsep{1pt}
  \caption{\small Evaluation of (a) clustering branch with different loss functions (i.e., \textbf{L$_1$}: L$_1$ distance, \textbf{L$_2$}: L$_2$ distance, and \textbf{KL}: KL-divergence) to measure the mismatch between two distributions and (b) mutual information estimated over input feature and different outputs (i.e., \textbf{CLS}: output of classification branch, \textbf{CLU}: output of clustering branch, and \textbf{CLS+CLU}: combined output of classification and clustering branches) on VisDA for open-set transfer.}
  \subtable[]{
  \begin{tabular}{c|ccc}
  \Xhline{2\arrayrulewidth}
  Method   &    Knwn    & Mean    & Overall  \\ \hline
  L$_1$    &    68.6    & 68.7    & 70.1     \\
  L$_2$    &    68.3    & 68.4    & 70.1     \\
  KL    &    70.4    & 70.5    & 71.6     \\
  \Xhline{2\arrayrulewidth}
  \end{tabular}
  \label{table:evaluationa}
  }
  \subtable[]{
  \begin{tabular}{c|ccc}
  \Xhline{2\arrayrulewidth}
  Method         &    Knwn    & Mean    & Overall   \\ \hline
  CLS        &    69.3    & 69.4    & 69.4      \\
  CLU        &    70.0    & 70.1    & 70.8      \\
  CLS+CLU    &    70.4    & 70.5    & 71.6      \\
  \Xhline{2\arrayrulewidth}
  \end{tabular}
  \label{table:evaluationb}
  }
  \vspace{-0.35in}
\end{table}

\textbf{Ablation Study.} Here we investigate how each design in our SE-CC influences the overall performance. Conditional Entropy (\textbf{CE}) incorporates an unsupervised conditional entropy loss into SE to drive the classifier's decision boundaries away from high-density target data regions in student model. KL-divergence Loss (\textbf{KL}) aligns the estimated cluster assignment distribution to the inherent cluster distribution for each target sample, targeting for refining feature to preserve the underlying structure of target domain. Mutual Information Maximization (\textbf{MIM}) further enhances the feature's suitability for downstream tasks by maximizing the mutual information among the input feature, the output classification and cluster assignment distributions. Table \ref{table:ablation} details the performance improvements on VisDA by considering different designs and their contributions for open-set domain adaptation in our SE-CC. CE is a general way to enhance classifier for target domain irrespective of any domain adaptation architectures. In our case, CE improves the Mean accuracy from 65.2\% to 66.3\%, which demonstrates that CE is an effective choice. KL and MIM are two specific designs in our SE-CC and the performance gain of each is 3.0\% and 1.2\% in Mean metric. In other words, our SE-CC leads to a large performance boost of 4.2\% in total in terms of Mean metric. The results verify the idea of exploiting underlying target data structure and mutual information maximization for open-set adaptation.

\textbf{Evaluation of Clustering Branch.} To study how the design of loss function in clustering branch affects the performance, we compare the use of KL-divergence in our SE-CC with L$_1$ and L$_2$ distance. The results in Table \ref{table:evaluationa} verify that KL-divergence is a better measure of mismatch between the classification and cluster assignment distributions than L$_1$ and L$_2$ distance, which yield inferior performance.

\textbf{Evaluation of Mutual Information Maximization.} Next, we evaluate different variants of MIM module in our SE-CC by estimating mutual information between input feature and different outputs, as shown in Table \ref{table:evaluationb}. CLS, CLU and CLS+CLU estimates the local and global mutual information between input feature and the output of classification branch, the output of clustering branch, and the combined output of two branches, respectively. Compared to our SE-CC without MIM module (Knwn: 69.3\%, Mean: 69.3\%, and Overall: 69.1\%), CLS and CLU slightly improves the performances by additionally exploiting the mutual information between input feature and the output of each branch. Furthermore, CLS+CLU obtains a larger performance boost, when combining the outputs from both branches for mutual information estimation. The results demonstrate the merit of exploiting the mutual information among the input feature and the combined outputs of two downstream tasks (i.e., classification and cluster assignment) in our MIM module.

\begin{figure}[!tb]
\vspace{-0.08in}
\centering {\includegraphics[width=0.5\textwidth]{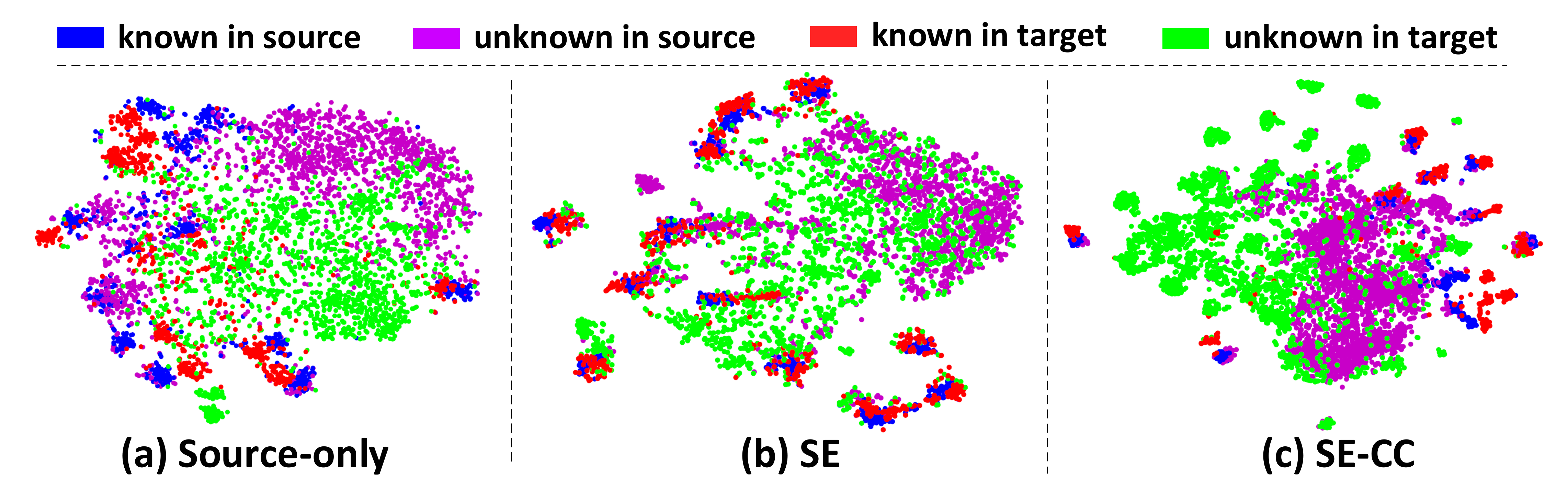}}
\vspace{-0.1in}
\caption{\small The t-SNE visualization of features learnt by (a) Source-only, (b) SE, and (c) SE-CC on VisDA for open-set adaptation.}
\label{fig:tsne}
\vspace{-0.2in}
\end{figure}

\textbf{Feature Visualization.}
We visualize the features learnt by Source-only, SE, and SE-CC with t-SNE \cite{maaten2008visualizing} on VisDA for open-set adaptation in Figure \ref{fig:tsne}(a)-(c). Compared to Source-only without domain adaptation, SE brings the two distributions of source and target closer, leading to domain-invariant representation. However, in SE, all target samples including unknown samples are enforced to match source samples, making it difficult to recognize unknown target samples with ambiguous semantics. Through the preservation of underlying target data structure for both known and unknown classes by SE-CC, the unknown target samples are separated from known target samples, and meanwhile the known samples in two domains are indistinguishable.

\section{Conclusion}
We have presented Self-Ensembling with Category-agnostic Clusters (SE-CC), which exploits the category-agnostic clusters in target domain for domain adaptation in both open-set and closed-set scenarios. Particularly, we study the problem from the viewpoint of how to separate unknown target samples from known ones and how to learn a hybrid network that nicely integrates category-agnostic clusters into Self-Ensembling. We initially perform clustering to decompose all target samples into a set of category-agnostic clusters. Next, an additional clustering branch is integrated into student model to align the estimated cluster assignment distribution to the inherent cluster distribution implicit in category-agnostic clusters. That enforces the learnt feature to preserve the underlying data structure in target domain. Moreover, the mutual information among the input feature, the outputs of classification and clustering branches is exploited to further enhance the learnt feature. Experiments conducted on Office and VisDA for both open-set and closed-set adaptation tasks verify our proposal. Performance improvements are observed when comparing to state-of-the-art techniques.

{\small
\bibliographystyle{ieee_fullname}
\bibliography{cvpr2020_conference}
}

\end{document}